\documentclass[english]{elsarticle}
\usepackage[T1]{fontenc}
\usepackage[latin9]{inputenc}
\usepackage{float}
\usepackage{url}
\usepackage{amsmath}
\usepackage{amssymb}
\usepackage{graphicx}
\usepackage{esint}

\makeatletter

\providecommand{\tabularnewline}{\\}
\floatstyle{ruled}
\newfloat{algorithm}{tbp}{loa}
\providecommand{\algorithmname}{Algorithm}
\floatname{algorithm}{\protect\algorithmname}

\usepackage{tikz}
\usetikzlibrary{bayesnet}

\def\tN{t\mathcal{N}}
\def\N{\mathcal{N}}

\def\tr{\mathrm{tr}}
\def\diag{\mathrm{diag}}
\def\vect{\mathrm{vec}}
\def\erf{\mathrm{erf}}
\def\G{\mathcal{G}}

\def\W{\mathcal{W}}
\def\d{\mathrm{d}}

\newcommand{\mbf}[1]{{\mathbf{#1}}}  

\newcommand{\ha}[1]{{\widehat{#1}}} 

\makeatother

\usepackage{babel}
\begin{document}
\title{Non-parametric Bayesian Models of Response Function in Dynamic Image Sequences}

\author{Ond\v{r}ej Tichý\fnref{myfootnote}} 
\author{Václav \v{S}mídl}
\address{Institute of Information Theory and Automation, Pod Vodárenskou V\v{e}\v{z}í 4, Prague 8, 18208, Czech Republic} 
\fntext[myfootnote]{corresponding author: otichy@utia.cas.cz, +420266052570}

\begin{abstract}
Estimation of response functions is an important task in dynamic medical imaging. This task arises for example in dynamic renal scintigraphy, where impulse response or retention functions are estimated, or in functional magnetic resonance imaging where hemodynamic response functions are required. These functions can not be observed directly and their estimation is complicated because the recorded images are subject to superposition of underlying signals. Therefore, the response functions are estimated via blind source separation and deconvolution. Performance of this algorithm heavily depends on the used models of the response functions. Response functions in real image sequences are rather complicated and finding a suitable parametric form is problematic. In this paper, we study estimation of the response functions using non-parametric Bayesian priors. These priors were designed to favor desirable properties of the functions, such as sparsity or smoothness. These assumptions are used within hierarchical priors of the blind source separation and deconvolution algorithm. Comparison of the resulting algorithms with these priors is performed on synthetic dataset as well as on real datasets from dynamic renal scintigraphy. It is shown that flexible non-parametric priors improve estimation of response functions in both cases. MATLAB implementation of the resulting algorithms is freely available for download.
\end{abstract}

\begin{keyword} 
Response function \sep Blind source separation \sep Dynamic medical imaging \sep Probabilistic models \sep Bayesian methods
\end{keyword}
\maketitle

\section{Introduction}

Computer analysis of dynamic image sequences offers an opportunity
to obtain information about organ function without invasive intervention.
A typical example is replacement of invasive blood sampling by computer
analysis of dynamic images \citep{lanz2014image}. The unknown input
function can be obtained by deconvolution of the organ time activity
curve and organ response function. Typically, both the input function
and the response functions are unknown. Moreover, the time-activity
curves are also not directly observed since the recorded images are
observed as superposition of multiple signals. The superposition arise
e.g. from partial volume effect in dynamic positron emission tomography
\citep{margadan2010ica} or dynamic and functional magnetic resonance
imaging \citep{chaari2012fast} or from projection of the volume into
planar dynamic scintigraphy \citep{di1982handling}. Analysis of the
dynamic image sequences thus require to separate the original sources
(source images) and their weights over the time forming the time-activity
curves (TACs). The TACs are then decomposed into input function and
response functions. Success of the procedure is dependent on the model
of the image sequence.

The common model for dynamic image sequences is the factor analysis
model \citep{martel2001extracting}, which assume linear combination
of the source images and TACs. Another common model is that TAC arise
as a convolution of common input function and source specific kernel
\citep{fleming1999comparison,taxt2012single}. The common input function
is typically the original signal from the blood and the role of convolution
kernels vary from application area: impulse response or retention
function in dynamic renal scintigraphy \citep{durand2008international}
or hemodynamic response function in functional magnetic resonance
imaging \citep{lindquist2009modeling}. In this paper, we will refer
to the source kernels as the response functions, however other interpretations
are also possible. 

Analysis of the dynamic image sequences can be done with supervision
of experienced physician or technician, who follows recommended guidelines
and uses medical knowledge. However, we aim at fully automated approach
where the analysis fully depends on the used model. The most sensitive
parameter of the analysis is the model of the response functions (i.e.
the convolution kernels). Many parametric models of response functions
has been proposed, including the exponential model \citep{chen2011tissue}
or piece-wise linear model \citep{kuruc1982idt,Tichy2014a}. An obvious
disadvantage of the approach is that the real response function may
differ from the assumed parametric models. Therefore, more flexible
class of models based on non-parametric ideas were proposed such as
averaging over region \citep{kershaw2000bayesian}, temporal regularization
using finite impulse response filters \citep{goutte2000modeling},
or free-form response functions using automatic relevance determination
principle in \citep{Tichy2014b}.

In this paper, we will study the probabilistic models of response
functions using Bayesian methodology within the general blind source
separation model \citep{miskin2000ensemble}. The Bayesian approach
was chosen for its inference flexibility and for its ability to incorporate
prior information of models \citep{woolrich2012bayesian,steinberg2014hierarchical}.
We will formulate the prior model for general blind source separation
problem with deconvolution \citep{Tichy2014b} where the hierarchical
structure of the model allow us to study various versions of prior
models of response functions. Specifically, we design different prior
models of the response functions with more parameters then the number
of points in the unknown response function. The challenge is to regularize
the estimation procedure such that all parameters are estimated from
the observed data. We will use the approximate Bayesian approach known
as the Variational Bayes method \citep{smidl2006vbm}. The resulting
algorithms are tested on synthetic as well as on real datasets.

\section{Probabilistic Blind Source Separation with Deconvolution\label{sec:BSSwithDC}}

In this Section, we introduce a model of dynamic image sequences.
Estimation of the model parameters yields an algorithm for Blind Source
Separation and Deconvolution. Prior models of all parameters except
for the response functions are described here while the priors for
the response functions will be studied in details in the next section.

\subsection{Model of Observation}

Each recorded image is stored as a column vector $\mathbf{d}_{t}\in\mathbf{R}^{p\times1},t=1,\dots,n$,
where $n$ is total number of recorded images. Each vector $\mathbf{d}_{t}$
is supposed to be an observation of a superposition of $r$ source
images $\mathbf{a}_{k}\in\mathbf{R}^{p\times1},k=1,\dots,r$, stored
again columnwise. The source images are weighted by their specific
activities in time $t$ denoted as $x_{1,t},\dots,x_{r,t}\equiv\overline{\mathbf{x}}_{t}\in\mathbf{R}^{1\times r}$.
Formally,
\begin{equation}
\mathbf{d}_{t}=\mathbf{a}_{1}x_{1,t}+\mathbf{a}_{2}x_{2,t}+\dots+\mathbf{a}_{r}x_{r,t}+\mathbf{e}_{t}=A\overline{\mathbf{x}}_{t}^{T}+\mathbf{e}_{t},\label{eq:dax}
\end{equation}
where $\mathbf{e}_{t}$ is the noise of the observation, $A\in\mathbf{R}^{p\times r}$
is the matrix composed from source images as its columns $A=[\mathbf{a}_{1},\dots,\mathbf{a}_{r}]$,
and symbol $()^{T}$ denotes transposition of a vector or a matrix
in the whole paper. The equation (\ref{eq:dax}) can be rewritten
in the matrix form. Suppose the observation matrix $D=[\mathbf{d}_{1},\dots,\mathbf{d}_{n}]\in\mathbf{R}^{p\times r}$
and the matrix with TACs in its columns, $X=[\overline{\mathbf{x}}_{1}^{T},\dots,\overline{\mathbf{x}}_{n}^{T}]^{T}\in\mathbf{R}^{n\times r}$.
Note that we will use the bar symbol, $\overline{\mathbf{x}}_{k}$,
to distinguish the $k$th row of matrix $X$, while $\mathbf{x}_{k}$
will be used to denote the $k$the column. Then, the (\ref{eq:dax})
can be rewritten into the matrix form as
\begin{equation}
D=AX^{T}+E.\label{eq:DAX}
\end{equation}

The tracer dynamics in each compartment is commonly described as convolution
of common input function, vector $\mathbf{b}\in\mathbf{R}^{n\times1}$,
and source specific response function (convolution kernel, mathematically),
vector $\mathbf{u}_{k}\in\mathbf{R}^{n\times1},k=1,\dots,r$ \citep{kuruc1982idt,diffey197699mtc,chen2011tissue}.
Using convolution assumption, each TAC can be rewritten as
\begin{equation}
\mathbf{x}_{k}=B\mathbf{u}_{k},\,\,\,\forall k=1,\dots,r,\label{eq:x_Bu}
\end{equation}
where the matrix $B\in\mathbf{R}^{n\times n}$ is composed from elements
of input function $\mathbf{b}$ as
\begin{equation}
B=\left(\begin{array}{cccc}
b_{1} & 0 & 0 & 0\\
b_{2} & b_{1} & 0 & 0\\
\dots & b_{2} & b_{1} & 0\\
b_{n} & \dots & b_{2} & b_{1}
\end{array}\right).\label{eq:B}
\end{equation}
Suppose the aggregation of response functions $U=[\mathbf{u}_{1},\dots,\mathbf{u}_{r}]\in\mathbf{R}^{n\times r}$.
Then, $X=BU$ and the model (\ref{eq:DAX}) can be rewritten as
\begin{equation}
D=AU^{T}B^{T}+E.\label{eq:AUBmodel}
\end{equation}

The task of subsequent analysis is to estimate the matrices $A$ and
$U$ and the vector $\mathbf{b}$ from the data matrix $D$.

\subsubsection{Noise Model}

We assume that the noise has homogeneous Gaussian distribution with
zero mean and unknown precision parameter $\omega$, $e_{i,j}=\N_{e_{i,j}}(0,\omega^{-1})$.
Then, the data model (\ref{eq:DAX}) can be rewritten as
\begin{equation}
f(D|A,X,\omega)=\prod_{t=1}^{n}\N_{\mathbf{d}_{t}}(A\overline{\mathbf{x}}_{t},\omega^{-1}I_{p}),\label{eq:fDAx}
\end{equation}
where symbol $\N$ denotes Gaussian distribution and $I_{p}$ is identity
matrix of the size given in its subscript. Since all unknown parameters
must have their prior distribution in the Variational Bayes methodology,
the precision parameter $\omega$ has a conjugate prior in the form
of the Gamma distribution
\begin{equation}
f(\omega)=\G_{\omega}(\vartheta_{0},\rho_{0}),\label{eq:fomega}
\end{equation}
with chosen constants $\vartheta_{0},\rho_{0}$.

\subsection{Probabilistic Model of Source Images\label{sub:Model-of-Source-Images}}

The only assumption on source images is that they are sparse, i.e.
only some pixels of source images are non-zeros. The sparsity is achieved
using prior model that favors sparse solution depending on data \citep{tipping2001sparse}.
We will employ the automatic relevance determination (ARD) principle
\citep{bishop2000variational} based on joint estimation of the parameter
of interest together with its unknown precision. Specifically, each
pixel $a_{i,j}$ of each source image has Gaussian prior truncated
to positive values (see Appendix \ref{sub:Truncated-Normal-Distribution})
with unknown precision parameter $\xi_{i,j}$ which is supposed to
have conjugate Gamma prior as
\begin{align}
f(a_{i,k}|\xi_{i,k})= & \tN_{a_{i,j}}(0,\xi_{i,k}^{-1}),\label{eq:fai}\\
f(\xi_{i,k})= & \G_{\xi_{i,j}}(\phi_{0},\psi_{0}),\label{eq:fxi}
\end{align}
for $\forall i=1,\dots,p,\forall k=1,\ldots,r,$ and $\phi_{0},\psi_{0}$
are chosen constants. The precisions $\xi_{i,j}$ form the matrix
$\Xi$ of the same size as $A$.

\subsection{Probabilistic Model of Input Function}

The input function $\mathbf{b}$ is assumed to be a positive vector;
hence, it will be modeled as truncated Gaussian distribution to positive
values with scaling parameter $\varsigma\in\mathbf{R}$ as
\begin{align}
f(\mathbf{b}|\varsigma)= & \tN(\mathbf{0}_{n,1},\varsigma^{-1}I_{n}),\label{eq:fb}\\
f(\varsigma)= & \G(\zeta_{0},\eta_{0}),\label{eq:fvarsigma}
\end{align}
where $\mathbf{0}_{n,1}$ denotes zeros matrix of the given size and
$\zeta_{0},\eta_{0}$ are chosen constants.

\subsection{Models of Response Functions}

\begin{figure}
\begin{centering}
\begin{tikzpicture}

\node[obs]                              (d) {$D$}; 
\node[latent, above=5mm of d, xshift=-2.0cm] (a) {$A$}; 
\node[latent, above=5mm of d]  (u) {$U$}; 
\node[latent, above=5mm of d, xshift=2.0cm]  (b) {$\mbf{b}$}; 
\node[latent, above=5mm of b] (vs) {$\varsigma$}; 
\node[latent, right=2.5cm of d]  (om) {$\omega$};
\node[latent, above=5mm of a] (xi) {$\Xi$};

\node[const, above=5mm of xi]  (hxi) {$\phi_0,\psi_0$}; 
\node[const, above=5mm of vs]  (hvs) {$\zeta_{0},\eta_{0}$}; 
\node[const, above=5mm of om]  (hom) {$\vartheta_0,\rho_0$};

\node[const, above=15mm of u] (hutext) {studied};
\node[const, above=10mm of u] (hutext2) {models};

\edge {a,u,b,om} {d} ; %
\edge {xi} {a} ;
\edge {hxi} {xi} ;
\edge {vs} {b} ; %
\edge {hom} {om} ; %
\edge {hvs} {vs} ; %

\plate {plu}{(u)(hutext)} {};
\end{tikzpicture} 
\par\end{centering}

\caption{\label{fig:hyerarchModel}Hierarchical model for blind source separation
with deconvolution problem.}
\end{figure}
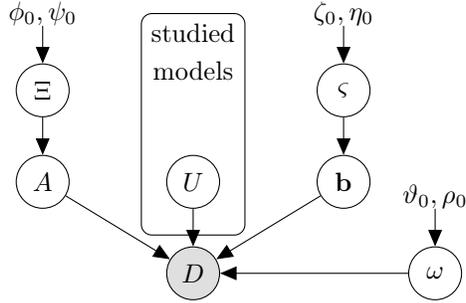

So far, we have formulated the prior models for source images $A$
and input function $\mathbf{b}$ from decomposition of the matrix
$D$. The task of this paper is to propose and study prior models
for response functions $U$ as illustrated in Figure \ref{fig:hyerarchModel}.
Different choices of the priors on the response functions have strong
influence on the results of the analysis which will be studied in
the next section.

\section{Non-parametric Prior Models of Response Function}

\begin{figure}
\begin{centering}
\includegraphics[width=0.8\textwidth]{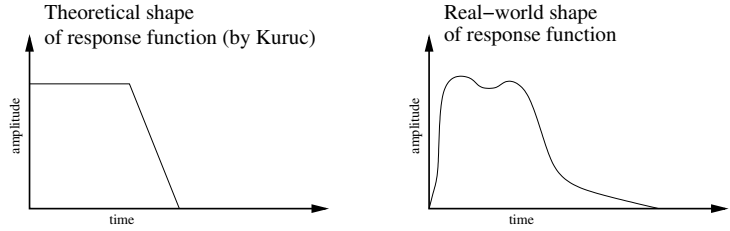}
\par\end{centering}

\caption{\label{fig:convol_example}Example of theoretical shape of response
function (by \citep{kuruc1982idt}), left, and corresponding real-world
shape of convolution kernels, right.}
\end{figure}

Here, we will formulate several prior models of response functions.
Our purpose is not to impose any parametric form as it was done, e.g.,
in \citep{kuruc1982idt,chen2011tissue} but model response function
as a free-form curve with only influence from their prior models.
The motivation is demonstrated in Figure \ref{fig:convol_example},
where a common parametric model \citep{kuruc1982idt} is compared
to an example of response function obtained from real data. While
the basic form of the response function is correct, exact parametric
form of the function would be very complex. Therefore, we prefer to
estimate each point on the response function individually. However,
this leads to overparametrization and poor estimates would result
without regularization. All models in this Section introduce regularization
of the non-parametric function via unknown covariance of the prior
with hyperparameters.

\subsection{Orthogonal Prior\label{sub:Relevances-on-Response}}

The first prior model assumes that each response function $\mathbf{u}_{k},k=1,\dots,r$,
is positive and each response function is weighted by its own precision
relevance parameter $\upsilon_{k}\in\mathbf{R}$ which has a conjugate
Gamma prior:
\begin{align}
f(\mathbf{u}_{k}|\upsilon_{k})= & \tN_{\mathbf{u}_{k}}\left(\mathbf{0}_{n,1},\upsilon_{k}^{-1}I_{n}\right),\\
f(\upsilon_{k})= & \G_{\upsilon_{k}}(\alpha_{0},\beta_{0}),
\end{align}
for $\forall k=1,\ldots,r$ and where $\alpha_{0},\beta_{0}$ are
chosen constants. 

The precision parameters $\upsilon_{k}$ serves for suppression of
weak response functions during iterative computation and therefore
as parameters responsible for estimation of number of relevant sources.

\subsection{Sparse Prior\label{sub:Sparse-Response-Functions}}

The model with sparse response functions has been introduced in \citep{Tichy2014b}.
The key assumption of this model is that the response functions are
most likely sparse which is modeled similarly as in case of source
images, Section \ref{sub:Model-of-Source-Images}, using the ARD principle.
Here, each element of response function $u_{k,j}$ has its relevance
parameter $\upsilon_{k,j}$ which is supposed to be conjugate Gamma
distributed. In vector notation, each response function $\mathbf{u}_{k}$
has its precision matrix $\Upsilon_{k}$ with precision parameters
$\upsilon_{k,j}$ on its diagonal and zeros otherwise. Then
\begin{align}
f(\mathbf{u}_{k}|\Upsilon_{k})= & \tN_{\mathbf{u}_{k}}\left(\mathbf{0}_{n,1},\Upsilon_{k}^{-1}\right),\\
f(\upsilon_{k,j})= & \G_{\upsilon_{k,j}}(\alpha_{0},\beta_{0}),\ \ \ \forall j=1,\ldots,n,\label{eq:fups_kj}
\end{align}
where $\alpha_{0},\beta_{0}$ are chosen constants.

The employed ARD principle should suppress the noisy parts of response
functions which should leads to clearer response functions and subsequently
to clearer TACs.

\subsection{Sparse Differences Prior\label{sub:Sparsity-on-Differences}}

Modeling of only sparsity in response functions could possibly leads
to arbitrary solution such as very non-smooth curve. The model of
differences in response functions allow us to formulate the model
favoring smooth response functions which is biologically reasonable
requirement. Let suppose the model of differences of response function
$\mathbf{u}_{k}$, $\nabla\mathbf{u}_{k}$, where the difference matrix
$\nabla$ is defined as
\begin{equation}
\nabla=\left(\begin{array}{cccc}
1 & -1 & 0 & 0\\
0 & 1 & \ddots & 0\\
0 & 0 & \ddots & -1\\
0 & 0 & 0 & 1
\end{array}\right),
\end{equation}
with ARD prior on each difference using precision parameter $\upsilon_{k,j}$
forming again precision matrix $\Upsilon_{k}$; however, with precisions
of differences on its diagonal. Then, we can formulate this problem
equally as
\begin{equation}
f(\nabla\mathbf{u}_{k}|\Upsilon_{k})=\tN_{\nabla\mathbf{u}_{k}}\left(\mathbf{0}_{n,1},\Upsilon_{k}^{-1}\right)\ \ \ \Longleftrightarrow\ \ \ f(\mathbf{u}_{k}|\Upsilon_{k})=\tN_{\mathbf{u}_{k}}\left(\mathbf{0}_{n,1},\nabla^{-1}\Upsilon_{k}^{-1}\nabla^{-T}\right),
\end{equation}
where symbol $()^{-T}$ denotes transpose and inversion of matrix.
The prior model is accompanied by prior model for precisions in the
same way as in (\ref{eq:fups_kj}):
\begin{equation}
f(u_{k,j})=\G_{\upsilon_{k,j}}(\alpha_{0},\beta_{0}),\ \ \ \forall j=1,\ldots,n,
\end{equation}
where $\alpha_{0},\beta_{0}$ are chosen constants.

\subsection{Wishart Prior\label{sub:Wishart-Covariance-Model}}

So far, we have modeled only the first or the second diagonal of the
precision matrix $\Upsilon_{k}$. Each of these approaches have its
advantages which we would like to generalize into estimation of several
diagonals of the prior covariance matrix. However, this is difficult
to solve analytically. Instead, we note that it is possible to create
the model for the full prior covariance matrix of the response functions
as well as their mutual interactions. For this task, we use vectorized
form of response functions denoted as $\mathbf{u}\in\mathbf{R}^{nr\times1}$,
$\mathbf{u}=\vect(U)=\left[\mathbf{u}{}_{1}^{T},\dots,\mathbf{u}{}_{r}^{T}\right]^{T}$.
This rearranging allow us to model mutual correlation between response
functions. The full covariance matrix $\Upsilon\in\mathbf{R}^{nr\times nr}$
can be modeled using Wishart distribution, see Appendix \ref{sub:Wishart-Distribution},
as
\begin{align}
f(\mathbf{u}|\Upsilon)= & \tN_{\mathbf{u}}\left(\mathbf{0}_{nr,1},\Upsilon^{-1}\right),\label{eq:Wish1}\\
f(\Upsilon)= & \W_{\Upsilon}\left(\alpha_{0}I_{nr},\beta_{0}\right),\label{eq:Wish2}
\end{align}
with scalar prior parameters $\alpha_{0},\beta_{0}$.

The advantage of this parametrization is obvious, the full covariance
matrix is estimated. The disadvantage is this model is that for estimation
$nr$ parameters in vector $\mathbf{u}$, we need to estimate $n^{2}r^{2}$
additional parameters in covariance structure. The problem is regularized
by the prior on $\Upsilon$, (\ref{eq:Wish2}), which is relatively
weak regularization with potential side effects. We try to suppress
these side effects in the next section.

\subsection{Wishart Prior with Localization\label{sub:Localized-Wishart-Covariance}}

\begin{figure}
\begin{centering}
\includegraphics[bb=40bp 30bp 310bp 302bp,clip,width=0.3\textwidth]{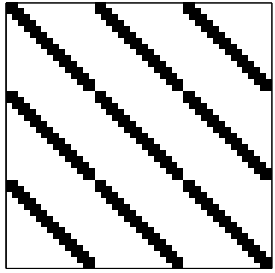}
\par\end{centering}

\caption{\label{fig:localMatrix}The used localization matrix $L$ for the
first two diagonals. The black pixels denote ones and the white pixels
denote zeros. This example is given for $n=15$ and $r=3$.}
\end{figure}
 Since restriction of the covariance structure to several diagonals
is infeasible in the considered dimensions, we apply an alternative
approach known as localization. This techniques originates in data
assimilation of atmospheric models \citep{hamill2001distance}. The
basic idea of the method is that the most information is localized
on the first two diagonals of the matrix $\Upsilon$ and its sub-matrices.
Hence, we can use Hadamard product, know also as element-wise product,
of the original estimates $\Upsilon$ and localization matrix $L$
of the same size as the matrix $\Upsilon$. The localization matrix
used in this paper for the first two diagonals is illustrated in Figure
\ref{fig:localMatrix}.

After localization, the model of response functions is the same as
in Section \ref{sub:Wishart-Covariance-Model}, (\ref{eq:Wish1})--(\ref{eq:Wish2}),
however, the estimate of $\Upsilon$, $\ha{\Upsilon}$, is replaced
by
\begin{equation}
\ha{\Upsilon}_{new}=\ha{\Upsilon}\circ L,
\end{equation}
where symbol $\circ$ denotes the Hadamard product. We will show that
this localization is a soft version of smoothing of the Wishart model
from Section \ref{sub:Wishart-Covariance-Model}; however, not so
strict as modeling of differences in Section \ref{sub:Sparsity-on-Differences}.

Theoretically, we could employ any conceivable localization as well
as smoother version of localization using smooth transitions between
ones and zeros; however, this is out of scope of this paper.

\subsection{Variational Bayes Approximate Solution}

The whole probabilistic model forming equations (\ref{eq:fDAx})--(\ref{eq:fomega}),
(\ref{eq:fai})--(\ref{eq:fvarsigma}), and selected response functions
model from Sections \ref{sub:Relevances-on-Response}--\ref{sub:Localized-Wishart-Covariance}.
The probabilistic model is solved using Variational Bayes (VB) method
\citep{smidl2006vbm}. Here, the solution is found in the form of
probability densities of the same type of the priors. The shaping
parameters of the posterior densities form a set of an implicit equations,
Appendix \ref{sec:Shaping-Parameters}, which is typically analytically
intractable and has to be solved iteratively.

\begin{algorithm}
\begin{enumerate}
\item Initialization:

\begin{enumerate}
\item Set prior parameters $\alpha_{0},\beta_{0},\vartheta_{0},\rho_{0},\phi_{0},\psi_{0},\zeta_{0},\eta_{0}$.
\item Set initial values for $\ha{A},\ha{A^{T}A},\ha{\Xi},\ha{\mathbf{u}},\ha{\mathbf{u}^{T}\mathbf{u}},\ha{\Upsilon},\ha{\mathbf{b}},\ha{\mathbf{b}^{T}\mathbf{b}},\ha{\varsigma},\ha{\omega}$.
\item Set the initial number of sources $r_{max}$.
\end{enumerate}
\item Iterate until convergence is reached using computation of shaping
parameters from Appendix \ref{sec:Shaping-Parameters}:

\begin{enumerate}
\item Source images $\mu_{\overline{\mathbf{a}}_{i}},\Sigma_{\overline{\mathbf{a}}_{i}}$
and their variances $\psi_{i},\phi_{i}$ $\forall i$ using (\ref{eq:img_ard1})--(\ref{eq:img_ard2}).
\item Response functions $\mu_{\mathbf{u}},\Sigma_{\mathbf{u}}$ and their
hyper-parameters depending on version of the prior: 

\begin{enumerate}
\item RelRF: (\ref{eq:mu_u_all}) and (\ref{eq:v1_1})--(\ref{eq:v1_2}), 
\item SparseRF: (\ref{eq:mu_u_all}) and (\ref{eq:v2_1})--(\ref{eq:v2_2}), 
\item SparDifRF: (\ref{eq:mu_u_all}) and (\ref{eq:v3_1})--(\ref{eq:v3_2}), 
\item WishRF: (\ref{eq:mu_u_all}) and (\ref{eq:v4_1})--(\ref{eq:v4_2}), 
\item LocWishRF: (\ref{eq:mu_u_all}) and (\ref{eq:v5_1})--(\ref{eq:v5_2}).
\end{enumerate}
\item Input function $\mu_{\mathbf{b}},\Sigma_{\mathbf{b}}$ and its variance
$\zeta,\eta$ using (\ref{eq:param_b1})--(\ref{eq:param_b2}).
\item Variance of noise $\vartheta,\rho$ using (\ref{eq:omega_1})--(\ref{eq:omega_2}).
\end{enumerate}
\item Report estimates of source images $\ha{A}$, response functions $\ha{U}$,
and input function $\ha{\mathbf{b}}$.
\end{enumerate}
\caption{\label{alg:SBSSDCvX}}
\end{algorithm}

The algorithms are summarized in Algorithm \ref{alg:SBSSDCvX}. All
prior parameters are set to $10^{-10}$ or $10^{+10}$ in order to
obtained non-informative priors. The initial response functions are
selected as pulses with different lengths with respect to cover the
typical structures while the initial input function is selected as
an exponential curve since the iterative solution could converge only
to a local minimum \citep{smidl2006vbm}.

\section{Experiments and Discussion}

We proposed five versions of model of non-parametric response functions
within the model of probabilistic blind source separation model in
Sections \ref{sub:Relevances-on-Response}--\ref{sub:Localized-Wishart-Covariance}.
The proposed algorithms are tested on simulated phantom study as well
as on representative clinical data set from dynamic renal scintigraphy.

\subsection{Synthetic Dataset}

\begin{figure}
\begin{centering}
\includegraphics[bb=90bp 60bp 880bp 660bp,clip,width=1\textwidth]{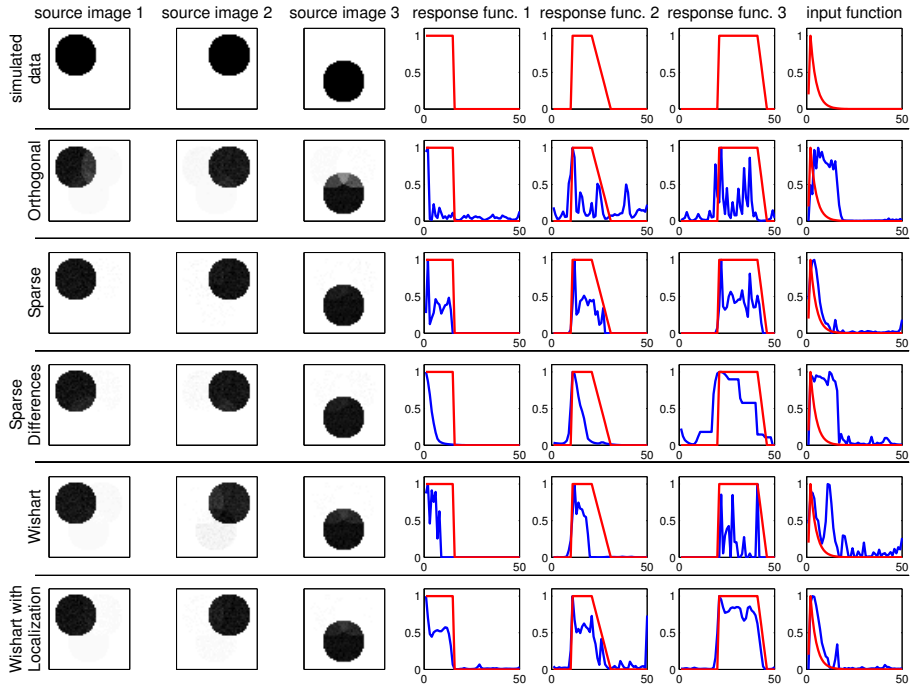}
\par\end{centering}

\caption{\label{fig:phantom_all}The results of the five studied methods on
synthetic dataset (the first row). The red lines are generated data
while the blue lines are estimated results from the respected methods.}
\end{figure}

Performance of the proposed models of response functions is first
studied on a synthetic dataset generated according to the model (\ref{eq:AUBmodel}).
The size of each image is $50\times50$ pixels and the number of simulated
time points is $n=50$. We simulate $3$ sources which are given in
Figure \ref{fig:phantom_all}, top row, using their source images
and response functions together with generated input function $\mathbf{b}$
(top row, right). We generate homogeneous Gaussian noise with standard
deviation $0.3$ of the signal strength.

The results of the five proposed models are given in Figure \ref{fig:phantom_all}
in the row-wise schema. Note that all algorithms are capable to estimate
the correct number of sources. It can be seen that all methods estimated
the source images correctly. The main differences are in estimated
response functions, the forth to the sixth columns, and estimated
input function, the seventh column. Note that only the first prior,
orthogonal, was not able to respect the sparse character of the modeled
response functions, all other priors were able to do so. The visual
results are accompanied by the corresponding mean square errors (MSE)
summarized in Table \ref{tab:mse_phantom}. Here, the MSE is computed
between the estimated response functions and their simulated values
as well as between the estimated input functions and its simulated
value for each method. The Wishart prior with localization outperforms
the other ones in estimation of response functions while it is comparable
with the sparse prior in estimation of input function.

\begin{table}
\begin{centering}
\begin{tabular}{|l|c|c|}
\hline 
Prior model of the response function  & total MSE on $U$ & MSE on \textbf{$\mathbf{b}$}\tabularnewline
\hline 
\hline 
Orthogonal, \hfill{}Sec. \ref{sub:Relevances-on-Response} & 34.50 & 6.58\tabularnewline
\hline 
Sparse, \hfill{}Sec. \ref{sub:Sparse-Response-Functions} & 18.76 & 0.99\tabularnewline
\hline 
Sparse Differences, \hfill{}Sec. \ref{sub:Sparsity-on-Differences} & 22.84 & 8.03\tabularnewline
\hline 
Wishart, \hfill{}Sec. \ref{sub:Wishart-Covariance-Model} & 30.33 & 3.43\tabularnewline
\hline 
Wishart with Localization, \hfill{}Sec. \ref{sub:Localized-Wishart-Covariance} & 9.26 & 1.19\tabularnewline
\hline 
\end{tabular}
\par\end{centering}

\caption{\label{tab:mse_phantom}Computed mean square errors (MSE) from the
simulated data.}
\end{table}

\subsubsection{Influence of Localization in Wishart Prior}

\begin{figure}
\begin{centering}
\includegraphics[width=0.8\textwidth]{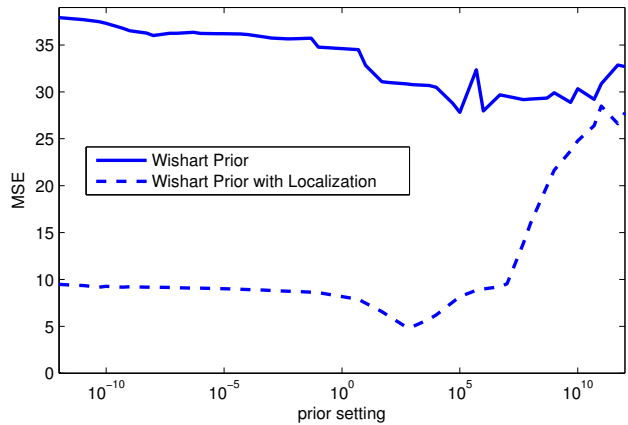}
\par\end{centering}

\caption{\label{fig:alpha0}Sensitivity study of MSE of estimated on the parameter
$\alpha_{0}$ of the Wishart prior and the Wishart prior with localization.}
\end{figure}

The effect of localization on the algorithm with the Wishart prior
is illustrated studied in Figure \ref{fig:alpha0} via sensitivity
study of the prior parameter $\alpha_{0}$ on the resulting MSE of
the response functions. The prior parameter $\beta_{0}$ is selected
as $10^{-10}$ for all cases. For large values of $\alpha_{0}$ the
results are comparable, however the localized version is improving
with decreasing $\alpha_{0}$. For values of $\alpha_{0}<1$, the
results of the localized version stabilize and become insensitive
to the exact value of $\alpha_{0}$.

\subsection{Datasets from Dynamic Renal Scintigraphy}

The methods from Sections \ref{sub:Relevances-on-Response}--\ref{sub:Localized-Wishart-Covariance}
were tested on real data from dynamic renal scintigraphy taken from
online database%
\footnote{Database of dynamic renal scintigraphy, \url{http://www.dynamicrenalstudy.org}
(accessed: 1st December 2014).%
}. We illustrate the possible outcome of the method on two distinct
datasets, numbers 84 and 42. Each dataset represent different behavior
of the methods.

Both sequences consist of $50$ frames taken after $10$ seconds and
both were preprocessed by selection region of the left kidney. The
data are expected to contain three sources of activity: (i) parenchyma,
the outer part of a kidney where the tracer is accumulated at the
first, (ii) pelvis, the inner part of a kidney where the accumulation
has physiological delay, and (iii) background tissues which is typically
active at the beginning of the sequence. Since the noise in scintigraphy
is Poisson distributed, the assumption of homogeneous Gaussian noise
(\ref{eq:fDAx}) can be achieved by asymptotic scaling known as the
correspondence analysis \citep{benali1993statistical} which transforms
the original data $D_{orig}$ as
\begin{equation}
d_{ij}=\frac{d_{ij,orig}}{\sqrt{\sum_{i=1}^{p}d_{ij,orig}\sum_{j=1}^{n}d_{ij,orig}}}.\label{eq:correspondence}
\end{equation}
\begin{figure}
\begin{centering}
\includegraphics[bb=85bp 55bp 860bp 640bp,clip,width=1\textwidth]{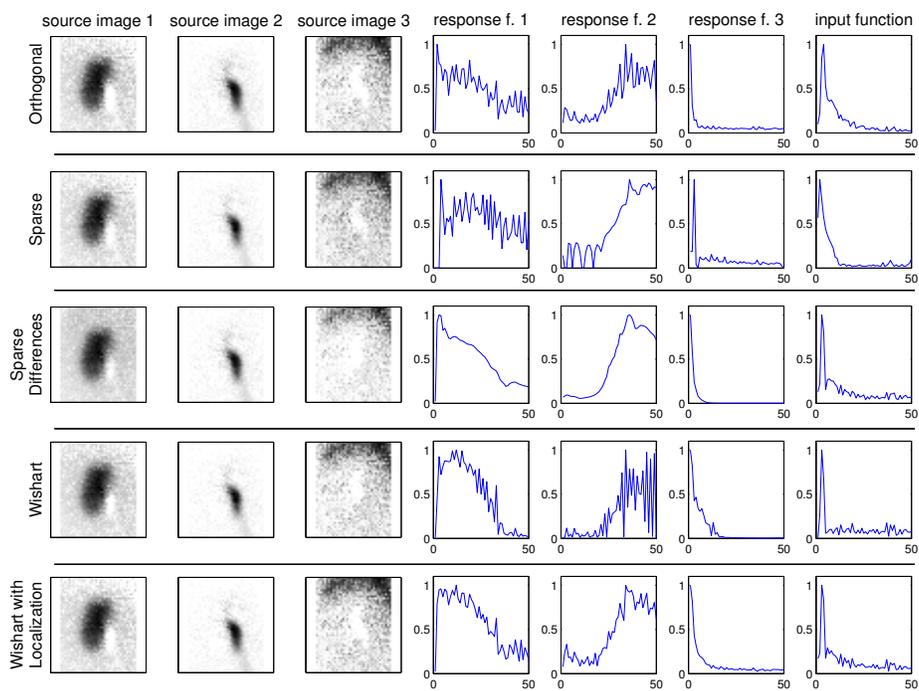}
\par\end{centering}

\caption{\label{fig:real_all_similar}Estimated source images (columns 1--3),
response functions (columns 4--6), and input functions (column 7)
using priors: Orthogonal, Sparse, Sparse Differences, Wishart, Wishart
with localization.}
\end{figure}

First, we applied the methods from Sections \ref{sub:Relevances-on-Response}--\ref{sub:Localized-Wishart-Covariance}
on dataset number $84$ as a typical non-controversial case. The results
are shown in Figure \ref{fig:real_all_similar} using the estimated
source images (columns 1--3), the estimated related response functions
(columns 4--6), and the estimated input function (column 7). The results
of all five methods are comparable with the main difference being
in the smoothness or non-smoothness of the estimated response functions.
This is most remarkable in the fifth column corresponding to the response
functions of the pelvis. The sparse prior prefers sparse solution
with many zeros, the sparse differences prior favors smooth solution
(i.e. many differences being equal to zeros), while the Wishart prior
models full covariance of response function where no smoothness is
incorporated. The hypothetical compromise of all versions seems to
be the Wishart prior with localization where the full covariance is
modeled and subsequently localized. However, the differences in this
case are relatively minor.

\begin{figure}
\begin{centering}
\includegraphics[bb=85bp 55bp 860bp 640bp,clip,width=1\textwidth]{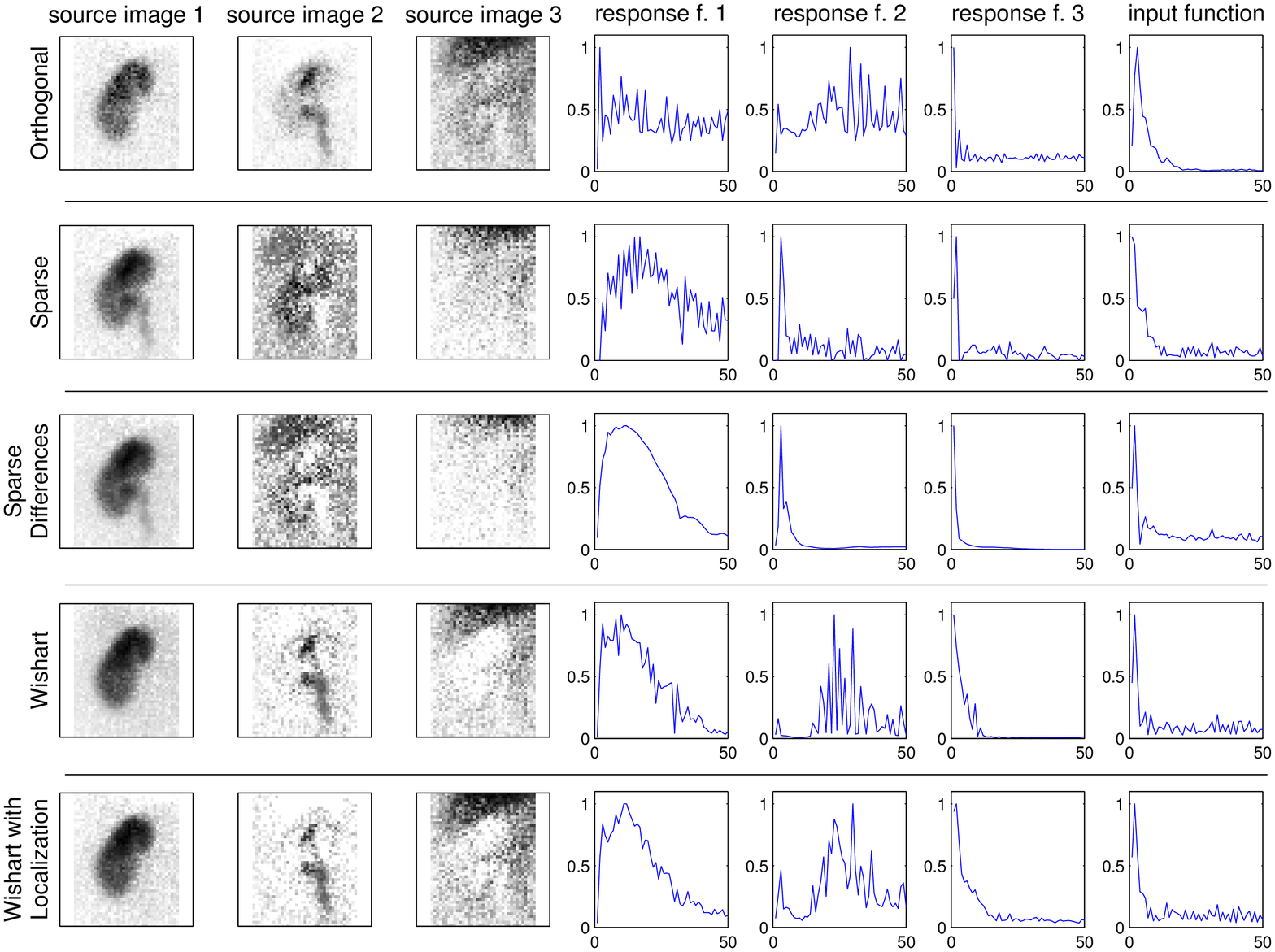}
\par\end{centering}

\caption{\label{fig:real_all_diff}Estimated source images (columns 1--3),
response functions (columns 4--6), and input functions (column 7)
using priors: Orthogonal, Sparse, Sparse Differences, Wishart, Wishart
with localization.}
\end{figure}

Second, we apply the methods \ref{sub:Relevances-on-Response}--\ref{sub:Localized-Wishart-Covariance}
on dataset number $42$ where different methods yield more distinct
results, see Figure \ref{fig:real_all_diff}. Note that the sparse
and the sparse differences priors were not able to separate the pelvis
which is mixed with the parenchyma in the first column while the orthogonal
prior estimated the source images reasonably; however, the response
functions of the parenchyma and the pelvis are clearly mixed. The
Wishart-based priors, Wishart and Wishart with localization, were
able to separate the parenchyma and the pelvis correctly together
with meaningful estimates of their response functions. The main difference
between the Wishart and the Wishart with localization priors is in
smoothness. The estimated response functions from the Wishart prior
with localization better matches the physiological expectations than
the estimates from the Wishart model. In this case, the use of more
complex prior models significantly outperform the simpler models. 

Indeed, the analysis of the full database would be of interest in
concrete application; however, it is not a goal of this paper.

\section{Conclusion}

A common model in functional analysis of dynamic image sequences assumes
that the observed images arise from superposition of the original
source images weighted by theirs time-activity curves. Each time-activity
curve is assumed to be a result of common input function and source-specific
response function, both unknown. Estimation of the model parameters
yields an algorithm for blind source separation and deconvolution.
The focus of this study is the prior model of the response functions
while the models of the source images and the input function are the
same. We propose five prior models of the response functions. The
first three prior models are based on automatic relevance determination
principle on the whole response functions, on each element of the
response function, and on the differences between elements of the
response functions, respectively. The forth model is based on full
model of covariance matrix using Wishart distribution while the fifth
model is based on the same prior; however, with additional localization
within the deconvolution algorithm. The advantage of all five models
is their flexibility in estimation of various shapes of response functions
since we do not impose any parametric form of them. The formulated
probabilistic models in the form of hierarchical priors are solved
using the Variational Bayes methodology.

The performance of the proposed methods is tested on simulated dataset
as well as on representative real datasets from dynamic renal scintigraphy.
It is shown that the behaviors of the methods well correspond with
their prior expectations. We conclude that the most complex model,
i.e. the Wishart model with localization, provide also the most desirable
results in the sense of mean square errors to the original simulated
data as well as in sense of biologically meaningfulness of the results
on the real datasets. Notably, the methods have no domain-specific
assumptions; hence, they can be used in other task in dynamic medical
imaging. The MATLAB implementation of all methods is available for
download in \url{http://www.utia.cz/bss_rf_priors/}.

\appendix

\section*{Acknowledgement}

This work was supported by the Czech Science Foundation, grant No.
GA13-29225S.

\section{Required Probability Distributions}

\subsection{Truncated Normal Distribution\label{sub:Truncated-Normal-Distribution}}

Truncated normal distribution, denoted as $\tN$, of a scalar variable
$x$ on interval $\left[a;b\right]$ is defined as
\begin{equation}
\tN(\mu,\sigma,\left[a,b\right])=\frac{\sqrt{2}\exp((x-\mu)^{2})}{\sqrt{\pi\sigma}(erf(\beta)-erf(\alpha))}\chi_{\left[a,b\right]}(x),
\end{equation}
where $\alpha=\frac{a-\mu}{\sqrt{2\sigma}}$, $\beta=\frac{b-\mu}{\sqrt{2\sigma}}$,
function $\chi_{\left[a,b\right]}(x)$ is a characteristic function
of interval $\left[a,b\right]$ defined as $\chi_{\left[a,b\right]}(x)=1$
if $x\in\left[a,b\right]$ and $\chi_{\left[a,b\right]}(x)=0$ otherwise.
$\erf()$ is the error function defined as $\erf(t)=\frac{2}{\sqrt{\pi}}\int_{0}^{t}e^{-u^{2}}\d u$. 

The moments of truncated normal distribution are
\begin{align}
\ha{x} & =\mu-\sqrt{\sigma}\frac{\sqrt{2}[\exp(-\beta^{2})-\exp(-\alpha^{2})]}{\sqrt{\pi}(\erf(\beta)-\erf(\alpha))},\label{eq:tN-x}\\
\ha{x^{2}} & =\sigma+\mu\widehat{x}-\sqrt{\sigma}\frac{\sqrt{2}[b\exp(-\beta^{2})-a\exp(-\alpha^{2})]}{\sqrt{\pi}(\erf(\beta)-\erf(\alpha))}.\label{eq:tN-x2}
\end{align}

\subsection{Wishart Distribution\label{sub:Wishart-Distribution}}

Wishart distribution $\W$ of the positive-definite matrix $X\in\mathbf{R}^{p\times p}$
is defined as
\begin{equation}
\W_{p}(\Sigma,\nu)=|X|^{\frac{\nu-p-1}{2}}2^{-\frac{\nu p}{2}}|\Sigma|^{-\frac{\nu}{2}}\Gamma_{p}^{-1}\left(\frac{\nu}{2}\right)\exp\left(-\frac{1}{2}\tr\left(\Sigma^{-1}X\right)\right),
\end{equation}
where $\Gamma_{p}\left(\frac{\nu}{2}\right)$ is the gamma function.
The required moment is:
\begin{align}
\widehat{X}= & \nu\Sigma.
\end{align}

\section{Shaping Parameters of Posteriors\label{sec:Shaping-Parameters}}

Shaping parameters of posterior distributions are given as:

\begin{align}
\Sigma_{\overline{\mathbf{a}}_{i}}= & \left(\ha{\omega}\sum_{j=1}^{n}(\ha{\overline{\mathbf{x}}_{j}^{T}\overline{\mathbf{x}}_{j}})+\diag(\ha{\Xi_{i}})\right)^{-1},\label{eq:img_ard1}\\
\mu_{\overline{\mathbf{a}}_{i}}= & \Sigma_{\overline{\mathbf{a}}_{i}}\ha{\omega}\sum_{j=1}^{n}(\ha{\overline{\mathbf{x}}_{j}}d_{i,j}),\\
\phi_{i}= & \phi_{i,0}+\frac{1}{2}\mathbf{1}_{r,1},\\
\psi_{i}= & \psi_{i,0}+\frac{1}{2}\diag\left(\ha{\overline{\mathbf{a}}_{i}^{T}\overline{\mathbf{a}}_{i}}\right),\label{eq:img_ard2}\\
\Sigma_{\mathbf{b}}= & \left(\ha{\varsigma}I_{n}+\ha{\omega}\sum_{i,j=1}^{r}(\ha{\overline{\mathbf{a}}_{i}^{T}\overline{\mathbf{a}}_{j}})\left(\sum_{k,l=0}^{n-1}\Delta_{k}^{T}\Delta_{l}\ha{u_{k+1,j}u_{l+1,i}}\right)\right)^{-1},\label{eq:param_b1}\\
\mu_{\mathbf{b}}= & \Sigma_{\mathbf{b}}\ha{\omega}\sum_{k=1}^{r}\left(\sum_{j=0}^{n-1}\Delta_{j}\ha{u_{j+1,k}}\right)^{T}D^{T}\ha{\mathbf{a}_{k}},\\
\zeta= & \zeta_{0}+\frac{n}{2},\ \ \eta=\eta_{0}+\frac{1}{2}\tr\left(\ha{\mathbf{b}^{T}\mathbf{b}}\right),\label{eq:param_b2}\\
\vartheta= & \vartheta_{0}+\frac{np}{2},\label{eq:omega_1}\\
\rho= & \rho_{0}+\frac{1}{2}\tr\left(DD^{T}-\ha{A}\ha{X}^{T}D^{T}-D\ha{X}\ha{A}^{T}\right)+\frac{1}{2}\tr\left(\ha{A^{T}A}\ha{X^{T}X}\right).\label{eq:omega_2}
\end{align}

Here, $\ha{x}$ denotes a moment of respective distribution, $\tr()$
denotes a trace of argument, $\diag()$ denotes a square matrix with
argument vector on diagonal and zeros otherwise or a vector composed
from diagonal element of argument matrix, $\mathbf{1}_{n,1}$ denotes
the matrix of ones of dimension $n\times1$, the auxiliary matrix
$\Delta_{k}\in\mathbb{\mathbf{R}}^{n\times n}$ is defined as $(\Delta_{k})_{i,j}=\begin{cases}
1, & \mathrm{if\,}i-j=k,\\
0, & \mathrm{otherwise,}
\end{cases}$, and standard moments of required probability distributions are given
Appendix \ref{sub:Truncated-Normal-Distribution} and \ref{sub:Wishart-Distribution}
and, e.g., in the appendix of \citep{smidl2006vbm}.

The shaping parameters for response functions are given in following
subsections while the parameter $\mu_{\mathbf{u}}$ is common for
all methods as
\begin{equation}
\mu_{\mathbf{u}}=\Sigma_{\mathbf{u}}\left(\ha{A^{T}A}\otimes\ha{\omega}\ha{B^{T}B}\right)\vect\left(\ha{B^{T}B}^{-1}\ha{B}^{T}D^{T}\ha{A}\ha{A^{T}A}^{-1}\right).\label{eq:mu_u_all}
\end{equation}

\subsection{Shaping Parameters for Orthogonal Prior}

\begin{align}
\Sigma_{\mathbf{u}}= & \left(\ha{A^{T}A}\otimes\ha{\omega}\ha{B^{T}B}+I_{n}\otimes\ha{\Upsilon}\right)^{-1},\label{eq:v1_1}\\
\alpha_{k}= & \alpha_{k,0}+\frac{n}{2},\ \ \beta_{k}=\beta_{k,0}+\frac{1}{2}\tr\left(\mbf{u}_{k}\mbf{u}_{k}^{T}\right),\label{eq:v1_2}
\end{align}

\subsection{Shaping Parameters for Sparse Prior}

\begin{align}
\Sigma_{\mathbf{u}}= & \left(\ha{A^{T}A}\otimes\ha{\omega}\ha{B^{T}B}+\diag(\vect(\ha{\Upsilon}))\right)^{-1},\label{eq:v2_1}\\
\alpha= & \alpha_{0}+\frac{1}{2}\mathbf{1}_{nr,1},\ \ \beta=\beta_{0}+\frac{1}{2}\diag\left(\ha{\mathbf{u}\mathbf{u}^{T}}\right),\label{eq:v2_2}
\end{align}

\subsection{Shaping Parameters for Sparse Differences Prior}

\begin{align}
\Sigma_{\mathbf{u}}= & \left(\ha{A^{T}A}\otimes\ha{\omega}\ha{B^{T}B}+(I_{r}\otimes\nabla)\ha{\Upsilon}(I_{r}\otimes\nabla^{T})\right)^{-1},\label{eq:v3_1}\\
\alpha_{k}= & \alpha_{k,0}+\mathbf{1}_{n,1}\frac{1}{2},\ \ \beta_{k}=\beta_{k,0}+\frac{1}{2}\diag\left(\nabla^{T}\mbf{u}_{k}\mbf{u}_{k}^{T}\nabla\right),\label{eq:v3_2}
\end{align}

\subsection{Shaping Parameters for Wishart Prior}

\begin{align}
\Sigma_{\mathbf{u}}= & \left(\ha{A^{T}A}\otimes\ha{\omega}\ha{B^{T}B}+\ha{\Upsilon}\right)^{-1},\label{eq:v4_1}\\
\Sigma_{\Upsilon}= & \left(\ha{\mathbf{u}\mathbf{u}^{T}}+(\alpha_{0}I_{nr})^{-1}\right)^{-1},\ \ \beta=\beta_{0}+1,\label{eq:v4_2}
\end{align}

\subsection{Shaping Parameters for Wishart Prior with Localization}

\begin{align}
\Sigma_{\mathbf{u}}= & \left(\ha{A^{T}A}\otimes\ha{\omega}\ha{B^{T}B}+\ha{\Upsilon}\circ L\right)^{-1},\label{eq:v5_1}\\
\Sigma_{\Upsilon}= & \left(\ha{\mathbf{u}\mathbf{u}^{T}}+(\alpha_{0}I_{nr})^{-1}\right)^{-1},\ \ \beta=\beta_{0}+1.\label{eq:v5_2}
\end{align}

\section*{References}

\bibliographystyle{elsarticle-num}
\bibliography{ot,ot_world}

\begin{thebibliography}{10}
\expandafter\ifx\csname url\endcsname\relax
  \def\url#1{\texttt{#1}}\fi
\expandafter\ifx\csname urlprefix\endcsname\relax\def\urlprefix{URL }\fi
\expandafter\ifx\csname href\endcsname\relax
  \def\href#1#2{#2} \def\path#1{#1}\fi

\bibitem{lanz2014image}
B.~Lanz, C.~Poitry-Yamate, R.~Gruetter, Image-derived input function from the
  vena cava for 18{F}-{FDG} {PET} studies in rats and mice, Journal of Nuclear
  Medicine 55~(8) (2014) 1380--1388.

\bibitem{margadan2010ica}
M.~Margad{\'a}n-M{\'e}ndez, A.~Juslin, S.~Nesterov, K.~Kalliokoski, J.~Knuuti,
  U.~Ruotsalainen, {ICA} based automatic segmentation of dynamic cardiac {PET}
  images., Information Technology in Biomedicine, IEEE Transactions on 14~(3)
  (2010) 795--802.

\bibitem{chaari2012fast}
L.~Chaari, T.~Vincent, F.~Forbes, M.~Dojat, P.~Ciuciu, Fast joint
  detection-estimation of evoked brain activity in event-related f{MRI} using a
  variational approach, Medical Imaging, IEEE Transactions on 32~(5) (2013)
  821--837.

\bibitem{di1982handling}
R.~Di~Paola, J.~Bazin, F.~Aubry, A.~Aurengo, F.~Cavailloles, J.~Herry, E.~Kahn,
  Handling of dynamic sequences in nuclear medicine, Nuclear Science, IEEE
  Transactions on 29~(4) (1982) 1310--1321.

\bibitem{martel2001extracting}
A.~Martel, A.~Moody, S.~Allder, G.~Delay, P.~Morgan, Extracting parametric
  images from dynamic contrast-enhanced mri studies of the brain using factor
  analysis, Medical image analysis 5~(1) (2001) 29--39.

\bibitem{fleming1999comparison}
J.~Fleming, P.~Kemp, {A comparison of deconvolution and the Patlak-Rutland plot
  in renography analysis}, Journal of Nuclear Medicine 40~(9) (1999) 1503.

\bibitem{taxt2012single}
T.~Taxt, R.~Jirik, C.~B. Rygh, R.~Gruner, M.~Bartos, E.~Andersen, F.-R. Curry,
  R.~K. Reed, Single-channel blind estimation of arterial input function and
  tissue impulse response in dce-mri, Biomedical Engineering, IEEE Transactions
  on 59~(4) (2012) 1012--1021.

\bibitem{durand2008international}
E.~Durand, M.~Blaufox, K.~Britton, O.~Carlsen, P.~Cosgriff, E.~Fine,
  J.~Fleming, C.~Nimmon, A.~Piepsz, A.~Prigent, et~al., {International
  Scientific Committee of Radionuclides in Nephrourology (ISCORN) consensus on
  renal transit time measurements}, in: Seminars in nuclear medicine, Vol.~38,
  Elsevier, 2008, pp. 82--102.

\bibitem{lindquist2009modeling}
M.~A. Lindquist, J.~Meng~Loh, L.~Y. Atlas, T.~D. Wager, Modeling the
  hemodynamic response function in fmri: efficiency, bias and mis-modeling,
  Neuroimage 45~(1) (2009) S187--S198.

\bibitem{chen2011tissue}
L.~Chen, P.~Choyke, T.-H. Chan, C.-Y. Chi, G.~Wang, Y.~Wang, Tissue-specific
  compartmental analysis for dynamic contrast-enhanced {MR} imaging of complex
  tumors, Medical Imaging, IEEE Transactions on 30~(12) (2011) 2044--2058.

\bibitem{kuruc1982idt}
A.~Kuruc, W.~Caldicott, S.~Treves, {Improved Deconvolution Technique for the
  Calculation of Renal Retention Functions.}, COMP. AND BIOMED. RES. 15~(1)
  (1982) 46--56.

\bibitem{Tichy2014a}
O.~Tich{\'{y}}, V.~{\v{S}}m\'{i}dl, M.~{\v{S}}{\'{a}}mal, Model-based
  extraction of input and organ functions in dynamic scintigraphic imaging,
  Computer Methods in Biomechanics and Biomedical Engineering: Imaging \&
  Visualization(in print, doi:10.1080/21681163.2014.916229).

\bibitem{kershaw2000bayesian}
J.~Kershaw, S.~Abe, K.~Kashikura, X.~Zhang, I.~Kanno, A bayesian approach to
  estimating the haemodynamic response function in event-related fmri,
  Neuroimage 11~(5) (2000) S474.

\bibitem{goutte2000modeling}
C.~Goutte, F.~A. Nielsen, L.~K. Hansen, Modeling the hemodynamic response in
  fmri using smooth fir filters, Medical Imaging, IEEE Transactions on 19~(12)
  (2000) 1188--1201.

\bibitem{Tichy2014b}
O.~Tich{\'{y}}, V.~{\v{S}}m\'{i}dl, Bayesian blind separation and deconvolution
  of dynamic image sequences using sparsity priors, Medical Imaging, IEEE
  Transaction on 34~(1) (2015) 1--9.

\bibitem{miskin2000ensemble}
J.~Miskin, {Ensemble learning for independent component analysis}, Ph.D.
  thesis, University of Cambridge (2000).

\bibitem{woolrich2012bayesian}
M.~W. Woolrich, Bayesian inference in fmri, NeuroImage 62~(2) (2012) 801--810.

\bibitem{steinberg2014hierarchical}
D.~M. Steinberg, O.~Pizarro, S.~B. Williams, Hierarchical bayesian models for
  unsupervised scene understanding, Computer Vision and Image Understanding(in
  print, http://dx.doi.org/10.1016/j.cviu.2014.06.004).

\bibitem{smidl2006vbm}
V.~\v{S}m{\'i}dl, A.~Quinn, {The Variational Bayes Method in Signal
  Processing}, Springer, 2006.

\bibitem{diffey197699mtc}
B.~Diffey, F.~Hall, J.~Corfield, {The 99mTc-DTPA dynamic renal scan with
  deconvolution analysis}, Journal of Nuclear Medicine 17~(5) (1976) 352.

\bibitem{tipping2001sparse}
M.~Tipping, Sparse {B}ayesian learning and the relevance vector machine, The
  journal of machine learning research 1 (2001) 211--244.

\bibitem{bishop2000variational}
C.~Bishop, M.~Tipping, Variational relevance vector machines, in: Proceedings
  of the 16th Conference on Uncertainty in Artificial Intelligence, 2000, pp.
  46--53.

\bibitem{hamill2001distance}
T.~Hamill, J.~Whitaker, C.~Snyder, Distance-dependent filtering of background
  error covariance estimates in an ensemble kalman filter, Monthly Weather
  Review 129~(11) (2001) 2776--2790.

\bibitem{benali1993statistical}
H.~Benali, I.~Buvat, F.~Frouin, J.~Bazin, R.~Paola, A statistical model for the
  determination of the optimal metric in factor analysis of medical image
  sequences ({FAMIS}), Physics in medicine and biology 38 (1993) 1065.

\end{thebibliography}

\end{document}